\documentclass[conference]{IEEEtran}
\IEEEoverridecommandlockouts

\usepackage{cite}
\usepackage{graphicx}
\PassOptionsToPackage{draft}{graphicx}
\usepackage{tikz}
\usepackage{amsmath,amssymb,amsfonts}
\usepackage[colorlinks=true, allcolors=blue]{hyperref}
\usepackage{algorithm}
\usepackage{enumitem}
\makeatletter
\RenewDocumentEnvironment{thebibliography}{m}{%
	\section*{\refname}%
	\list{\@biblabel{\@arabic\c@enumiv}}{%
		\itemsep=0pt
		\parsep=0pt
		\topsep=0pt
		\small
		\settowidth\labelwidth{\@biblabel{#1}}%
		\leftmargin\labelwidth
		\advance\leftmargin\labelsep
		\@openbib@code
		\usecounter{enumiv}%
		\let\p@enumiv\@empty
		\renewcommand\theenumiv{\@arabic\c@enumiv}
	}
	\sloppy
	\clubpenalty4000
	\@clubpenalty \clubpenalty
	\widowpenalty4000
	\sfcode`\.\@m
}{%
	\def\@noitemerr{\@latex@warning{Empty `thebibliography' environment}}%
	\endlist
}
\makeatother
\usepackage{algorithmic}
\usepackage{textcomp}
\usepackage{xcolor}
\usepackage{multirow}
\usepackage{booktabs}
\def\BibTeX{{\rm B\kern-.05em{\sc i\kern-.025em b}\kern-.08em
    T\kern-.1667em\lower.7ex\hbox{E}\kern-.125emX}}
\usepackage{eso-pic}

\AddToShipoutPictureBG*{%
	\AtPageUpperLeft{%
		\put(60,-30){%
			\parbox{0.8\paperwidth}{\centering
				\footnotesize
				This paper has been accepted for presentation at the IEEE OCEANS 2025 Brest conference.
				Please cite the paper as:\\J. Su, et al. "A Convex and Global Solution for the P$n$P Problem in 2D Forward-Looking Sonar" in OCEANS 2025-Brest. IEEE, 2025, pp. 1-7.\\
			}
		}
	}
}

\begin{document}

\title{\bf A Convex and Global Solution for the P$n$P Problem in 2D Forward-Looking Sonar
}

\author{Jiayi Su$^{1}$, Jingyu Qian$^{1}$, Liuqing Yang$^{1,4}$,\\ Yufan Yuan$^{2}$, Yanbing Fu$^{2,3}$, Jie Wu$^{2}$, Yan Wei$^{2}$ and Fengzhong Qu$^{2,3}$
\thanks{This work is in part supported by Natural Science Foundation of China Project \#U23A20339, and Guangdong Provincial Project \#2023ZDZX1037 and \#2023ZT10X009.}
\thanks{$^{1}$Jiayi Su, Jingyu Qian and Liuqing Yang are with the Intelligent Transportation Thrust, The Hong Kong University of Science and Technology (Guangzhou), Guangzhou 511458, China. Emails: \{jiayis, jingyuqian\}@hkust-gz.edu.cn, lqyang@ust.hk}
\thanks{$^{2}$Yufan Yuan, Yanbing Fu, Jie Wu, Yan Wei and Fengzhong Qu are with 1) the Engineering Research Center of Oceanic Sensing Technology and Equipment, Ministry of Education, Ocean College, Zhejiang University, Zhoushan 316021, China. and 2) the Provincial Key Laboratory of Cutting-edge Scientific Instruments R\&D and Application, Ocean College, Zhejiang University, Zhoushan 316021, China. Emails: \{y\_yuf, kikorfuyb, Coral\_WJ, redwine447, jimqufz\}@zju.edu.cn}
\thanks{$^{3}$Yanbing Fu and Fengzhong Qu are also with the Hainan Institute of Zhejiang University, Sanya 572025, China.}
\thanks{$^{4}$Liuqing Yang is also with the Intelligent Transportation Thrust, The Hong Kong University of Science and Technology (Guangzhou), Guangzhou 511458, China, and with the Department of Electronic and Computer Engineering, The Hong Kong University of Science and Technology, Hong Kong SAR 999077, China.}
\thanks{Liuqing Yang is the corresponding author.}
}

\maketitle

\begin{abstract}
	The perspective-$n$-point (P$n$P) problem is important for robotic pose estimation. It is well studied for optical cameras, but research is lacking for 2D forward-looking sonar (FLS) in underwater scenarios due to the vastly different imaging principles. In this paper, we demonstrate that, despite the nonlinearity inherent in sonar image formation, the P$n$P problem for 2D FLS can still be effectively addressed within a point-to-line (PtL) 3D registration paradigm through orthographic approximation. The registration is then resolved by a duality-based optimal solver, ensuring the global optimality. For coplanar cases, a null space analysis is conducted to retrieve the solutions from the dual formulation, enabling the methods to be applied to more general cases. Extensive simulations have been conducted to systematically evaluate the performance under different settings. Compared to non-reprojection-optimized state-of-the-art (SOTA) methods, the proposed approach achieves significantly higher precision. When both methods are optimized, ours demonstrates comparable or slightly superior precision.
\end{abstract}

\begin{IEEEkeywords}
	perspective-$n$-point (P$n$P) problem, forward-looking sonar (FLS), point-to-line registration.
\end{IEEEkeywords}

\section{Introduction}
\textbf{Background.} Pose estimation using exteroceptive sensors is crucial for achieving autonomy in underwater robotics. Common exteroceptive sensors such as optical cameras are cheap and can be easily integrated but their perceptional range is highly sensitive to the turbidity of the water. On the other hand, the 2D forward-looking sonar (FLS) has gradually gained popularity due to the insensitivity of acoustic waves to turbidity, making it a complementary modality for perception systems and, in severe situations, the only effective sensor. Therefore, in this paper, we focus on addressing the pose estimation problem for 2D FLS, specifically the perspective-$n$-point (P$n$P) problem, which is defined as follows: Given the correspondences between the 3D points in world coordinates and their 2D observations, determine the transformation matrix between sensor and world coordinates. The P$n$P problem is ubiquitous in robotics applications, such as simultaneous localization and mapping (SLAM), augmented reality (AR), and structure from motion (SfM). While it has been extensively studied in the camera community, the development and understanding in the context of 2D FLS remain insufficient. 

\textbf{Related Works.} In \cite{negahdaripour2005calibration}, the intrinsic parameters and pose of the 2D FLS relative to a planar grid-like target were estimated alternately. The pose estimation relied on an iterative method, which required a good initial guess. Due to the problem's highly non-convex nature, this approach is susceptible to local minima. \cite{brahim20113d} employed the covariance matrix adaptation evolution strategy (CMA-ES) to iteratively estimate the 2D FLS pose parameters relative to nonplanar targets. This approach also requires a good initial guess. By first acquiring a closed-form solution and then refining it through iterative optimization, \cite{wang2020planar,wang2020acmarker,wang2024acoustic} can obtain stable and accurate pose estimation results. However, determining the translation along the $z$-axis also relies on iterative refinement, increasing the computational cost. \cite{yang2020extrinsic} can solve $t_z$ analytically, but it uses single point and easily encounters singular situations when noise level is high. A unique $t_z$ estimation can then be retrieved from plane fitting, however, this requires coplanar point configuration. Most recently, \cite{sheng2024bestanp} reported solving the P$n$P problem in 2D FLS through a bi-step paradigm: first estimating $\mathbf{t}$ using only range measurements, then determining orientation based on measured bearing angles and the estimated $\hat{\mathbf{t}}$. Although this method is fully analytical and highly efficient, it requires nonplanar 3D point configurations. Another class of P$n$P-like problems involves using a known model instead of just point information to estimate the sonar's poses \cite{park2022robust,wang2022simulator}. Although these methods are more robust, they require a thorough prior understanding of the scene.

\textbf{Contributions.} To take it a step further, in this paper, we study the P$n$P problem in 2D FLS and propose a convex and global solution. The contributions are summarized as follows:
\begin{enumerate}
	\item Despite the nonlinearity inherent in 2D FLS image formation, we demonstrate that the P$n$P problem can still be effectively addressed within a point-to-line (PtL) 3D registration paradigm using orthographic approximation.
	\item A duality-based optimal solver is applied to this registration problem, incorporating an additional null space analysis to address the degeneracy arising in coplanar cases of 2D FLS.
	\item Through extensive simulations, we demonstrate that the proposed method achieves significantly higher precision than non-reprojection-optimized state-of-the-art (SOTA) approaches. When both methods are optimized, ours demonstrates comparable or slightly superior precision.
\end{enumerate}

\section{Preliminaries}
\label{sec:Preliminaries}
\subsection{Projection Model of 2D FLS}
\label{subsec:projection}
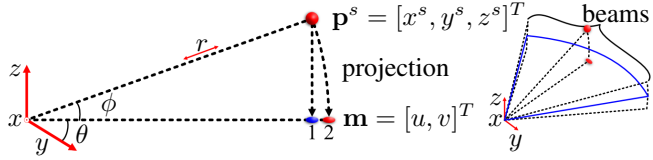
\begin{figure}[!t]
	\centering
	\begin{tikzpicture}
		\node[anchor=south west,inner sep=0] (image) at (0,0) 
		{\includegraphics[width=\linewidth]{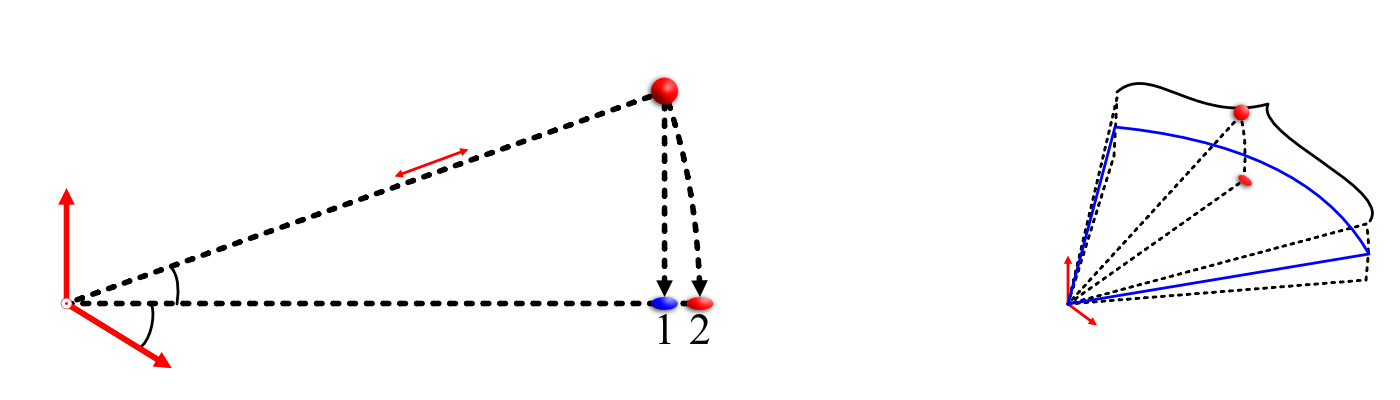}};
		\begin{scope}[shift={(image.south west)}, x={(image.south east)}, y={(image.north west)}]
			\node at (0.4, 1) {1: orthgraphic (approximated); 2: arc projection};
			\node at (0.03,0.24) {$x$};
			\node at (0.03,0.5) {$z$};
			\node at (0.07,0.1) {$y$};
			\node at (0.75,0.24) {$x$};
			\node at (0.75,0.35) {$z$};
			\node at (0.78,0.12) {$y$};
			\node at (0.31,0.63) {$r$};
			\node at (0.6,0.5) {projection};
			\node at (0.93,0.8) {beams};
			\node at (0.17,0.32) {$\phi$};
			\node at (0.13,0.17) {$\theta$};
			\node at (0.65,0.78) {$\mathbf{p}^s = [x^s, y^s, z^s]^T$};
			\node at (0.62,0.25) {$\mathbf{m} = [u, v]^T$};			
		\end{scope}
	\end{tikzpicture}
	\caption{The projection model of 2D FLS. The left part is a side view of one of the dashed sectors in the right part. The arc projection represents the true working mechanism of the sonar.}
	\label{fig:sonar_sketch}
\end{figure}

For the P$n$P problem in 2D FLS, we focus on the geometric projection characteristics. For details on its working mechanism, please refer to \cite{belcher2002dual}. A 3D point $\mathbf{p}^s$ in sonar coordinate can be described using $\left(r, \theta, \phi\right)$ in spherical coordinates:
\begin{equation}
	\mathbf{p}^s = 
	\begin{bmatrix}
		x^s \\
		y^s \\
		z^s
	\end{bmatrix} = 
	\begin{bmatrix}
		r\cos\phi\sin\theta \\
		r\cos\phi\cos\theta \\
		r\sin\phi
	\end{bmatrix},
	\label{equ:spherical-to-cartesian}
\end{equation}
where $r$ is the measured distance between the  $\mathbf{p}^s$ and the center of the sonar's transmitting array, $\theta$ is the bearing angle, $\phi$ is the elevation angle. During measurements, $\phi$ is lost, causing \textit{elevation ambiguity}. Thus, the corresponding projected 2D point $\mathbf{m}$ can be obtained through
\begin{equation}
	\mathbf{m} = 
	\begin{bmatrix}
		u \\ v 
	\end{bmatrix} = 
	\begin{bmatrix}
		r\sin\theta \\
		r\cos\theta
	\end{bmatrix}.
	\label{equ:spherical-to-pixel}
\end{equation}
For each point $i$, combining \eqref{equ:spherical-to-cartesian} and \eqref{equ:spherical-to-pixel}, we get
\begin{equation}
	\begin{bmatrix}
		u_i \\
		v_i 
	\end{bmatrix} =
	\begin{bmatrix}
		\cos^{-1}\phi_i & 0 & 0 \\
		0 & \cos^{-1}\phi_i & 0
	\end{bmatrix}
	\begin{bmatrix}
		x^s_i \\ y^s_i \\ z^s_i
	\end{bmatrix}.
	\label{equ:cartesian-to-pixel}
\end{equation}
For sonars from different manufacturers, the typical range of $\phi$ values is approximately $12^\circ$ to $30^\circ$ \footnote{\href{https://www.blueprintsubsea.com/downloads/oculus/DA-148-P01443-09.pdf\#page=2.00}{Blueprint Subsea Oculus M1200d} and \href{https://www.kongsbergdiscovery.online/m3_sonar/ref/m3_sonar_ref_en_us_lores.pdf\#page=225.57}{Kongsberg M3 sonar.}}, which results in $\cos\phi \in [0.9659,1]$. For some problems in the fields of computer vision and robotics, researchers have linearized the term $\cos\phi$ to simplify the analysis process by assuming $\cos\phi = \alpha$. If $\alpha = 1$, the projection is approximated as an orthographic model\cite{su2024analysis, hurtos2015fourier}. The projection characteristics are concluded in Fig. \ref{fig:sonar_sketch}.

\subsection{Problem Formulation}
The relationship between 3D points in world coordinate $\mathbf{p}^w$ and the corresponding measurements $\mathbf{m}$ can be described as follows:
\begin{equation}
	\underbrace{
		\begin{bmatrix}
			u_i \\ v_i \\ 1 
	\end{bmatrix}}_{\mathbf{m}_i} =
	\underbrace{
		\begin{bmatrix}
			\cos^{-1}\phi_i & 0 & 0 & 0 \\
			0 & \cos^{-1}\phi_i & 0 & 0 \\
			0 & 0 & 0 & 1
	\end{bmatrix}}_{\mathbf{M}_p}
	\underbrace{
		\begin{bmatrix}
			\mathbf{R} & \mathbf{t} \\
			\mathbf{0} & 1
	\end{bmatrix}}_{\mathbf{T}^s_w}
	\underbrace{
		\begin{bmatrix}
			x^w_i \\ y^w_i \\ z^w_i \\ 1
	\end{bmatrix}}_{\mathbf{p}^w_i},
	\label{equ:complete-relationship}
\end{equation}
where $\mathbf{R} \in SO(3)$ and $\mathbf{t} \in \mathbb{R}^{3}$ are the orientation and position to be determined, $\mathbf{T}^s_w$ is composed of $\mathbf{R}$ and $\mathbf{t}$ to represent the homogeneous coordinate transformation from world to sonar frame, $\mathbf{M}_p$ represents the projection function(or matrix). We simplify the use of (in)homogeneous coordinates, as readers can directly know which one is used without effort. To solve for $\mathbf{R}$ and $\mathbf{t}$, a common approach is to aggregate each pair $\{\left(\mathbf{p}^w_i, \mathbf{m}_i\right)\}_{i=1}^N$ into least-square form:
\begin{equation}
	\begin{aligned}[t]
		\mathop{\arg\min}\limits_{\mathbf{R},\mathbf{t}}
		\sum_{i=1}^{N} \| \mathbf{M}_p\mathbf{T}_w^s \mathbf{p}^w_i - \mathbf{m}_i \|^2,&
		\\	
		\text{s.t.} \qquad \forall i, \quad \phi_\text{min} \le \phi_i \le \phi_\text{max},&
		\\
		\phi_i = \frac{\sqrt{(x^s_i)^2 + (y^s_i)^2}}{\sqrt{(x^s_i)^2+(y^s_i)^2+(z^s_i)^2}}.&
	\end{aligned}
	\label{equ:general-error-func}
\end{equation}
Due to the non-convexity and nonlinearity in \eqref{equ:general-error-func}, gradient-based iterative optimization methods, such as Gauss-Newton or Levenberg–Marquardt, are prone to getting trapped in local minima\cite{hartley2003multiple}. Some efforts have been made to obtain an initial guess, which is then followed by a minimization on reprojection residual \cite{wang2024acoustic,yang2020extrinsic}. While the precision improvement through optimization is significant, the optimization itself is computationally expensive, hindering its use in time-sensitive applications. In this work, we demonstrate that when solving the P$n$P problem in 2D FLS within a PtL 3D registration paradigm, the initial guess can be sufficiently accurate without requiring further optimization.

\section{Methods}
\subsection{From P$n$P to PtL 3D Registration.}

\begin{figure}[!t]
	\centering
	\begin{tikzpicture}
		\node[anchor=south west,inner sep=0] (image) at (0,0) 
		{\includegraphics[width=\linewidth]{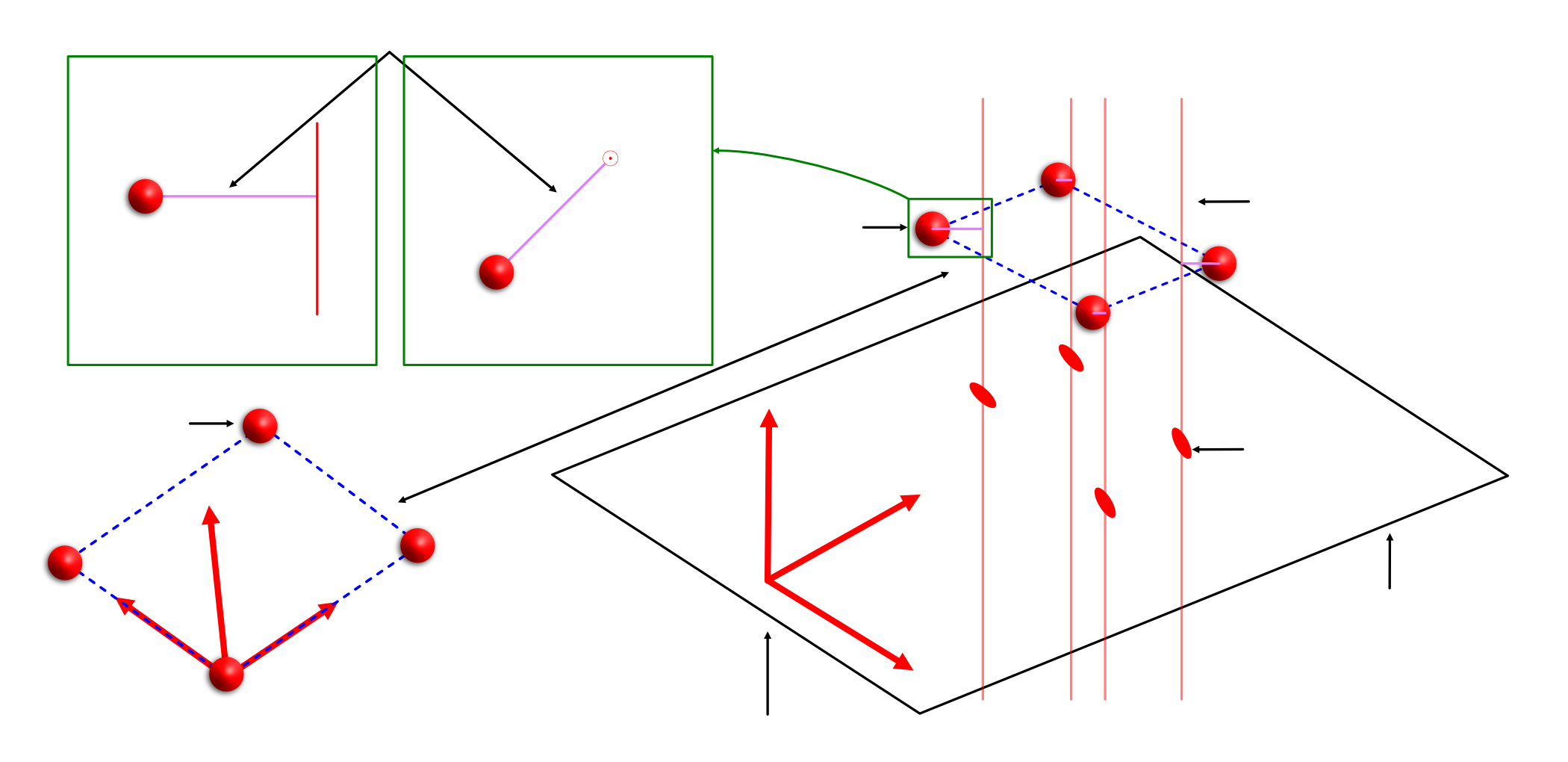}};
		\begin{scope}[shift={(image.south west)}, x={(image.south east)}, y={(image.north west)}]
			\node at (0.25, 0.97) {point to line};
			\node at (0.13, 0.56) {side view};
			\node at (0.33, 0.56) {top view};
			\node at (0.1,0.45) {$\mathbf{p}^w$};
			\node at (0.53,0.7) {$\mathbf{p}^s$};
			\node at (0.36,0.47) {$\mathbf{T}$};
			\node at (0.22,0.16) {$x$};
			\node at (0.15,0.3) {$z$};
			\node at (0.07,0.16) {$y$};
			\node at (0.15,0.07) {world coordinate};
			\node at (0.5,0.04) {sonar coordinate};
			\node at (0.88,0.17) {imaging plane};
			\node at (0.84,0.41) {$\mathbf{m}/\mathbf{o}$};
			\node at (0.59,0.16) {$x$};
			\node at (0.51,0.44) {$z$};
			\node at (0.55,0.35) {$y$};
			\node at (0.53,0.82) {enlarge};
			\node at (0.88,0.74) {$\mathbf{d} = \begin{bmatrix}
					0 \\ 0 \\ 1
				\end{bmatrix}$};
		\end{scope}
	\end{tikzpicture}
	\caption{The illustration of the point-to-line cost used in our method. 4 points of a rectangle are given in world coordinate, then transformed into sonar coordinate and projected as 4 pixels. Lines with direction $\mathbf{d} = [0 \ 0 \ 1]^T$ are shown as red lines passing through $\mathbf{m}/\mathbf{o}$. The transformed $\mathbf{p}^w$, which is $\mathbf{p}^s$, has its distance to the lines represented by purple line segments. The top-left green boxes are enlarged side/top-views.}
	\label{fig:point-to-line}
\end{figure}
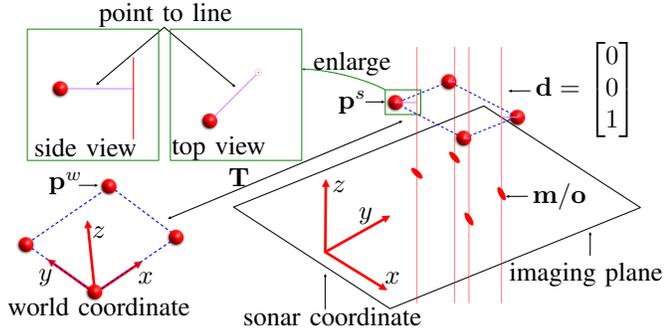

Recalling \eqref{equ:complete-relationship}, by approximating $\cos^{-1}\phi$ as $1$ through the orthographic model, we have
\begin{equation}
	\mathbf{m} = \mathbf{M}_p(\mathbf{Rp}^w+\mathbf{t}),
	\quad
	\mathbf{M}_p = 
	\begin{bmatrix}
		1 & 0 & 0 \\ 0 & 1 & 0
	\end{bmatrix}.
\end{equation}
Then arrange it into the form of least squares, we have
\begin{equation}
	\begin{split}
		\| \mathbf{M}_p(\mathbf{Rp}^w+\mathbf{t}) - \mathbf{m} \|^2
		= \| \mathbf{M}_p(\mathbf{Rp}^w+\mathbf{t}) - 
		\mathbf{M}_p\underbrace{
			\begin{bmatrix}
				\mathbf{m} \\ 0
		\end{bmatrix}}_{\mathbf{o}}
		\|^2
		\\ 
		=\| \mathbf{Rp}^w+\mathbf{t} - \mathbf{o}
		\|_{\mathbf{M}_p^T\mathbf{M}_p}^2 
		= \| \mathbf{R}\mathbf{p}^w+\mathbf{t}-\mathbf{o} \|^2_{\mathbf{I}_3 - \mathbf{dd}^T}.
	\end{split}
	\label{equ:cost}
\end{equation}
where $\mathbf{o}$ is a point lying on the $z=0$ plane in sonar coordinate, and $\mathbf{d}= [0 \ 0 \ 1]^T$. What we obtain is a standard PtL residual, with the geometric illustration provided in Fig. \ref{fig:point-to-line}.
We adopt the solver proposed by \cite{briales2017convex} to solve the PtL 3D registration problem. The reason for this choice is that it is time-tested and effective in the 2D FLS context. From the output of the solver, we obtain $\mathbf{R}$ and $\mathbf{t}_{xy}$. For $t_z$, since this dimension is aligned with the orthographic projection direction, the solver only returns $0$. Further effort should be made to retrieve $t_z$.

\subsection{Solving for $t_z$}
For each $i$, we have
\begin{equation}
	r_i = \sqrt{(x_i^s)^2 + (y_i^s)^2 + (z_i^s)^2} = \sqrt{(u_i)^2 + (v_i)^2}.
	\label{equ:r-constraint}
\end{equation}
By substituting $\mathbf{R}$, $\mathbf{p}^w_i$ and $\mathbf{m}$ into \eqref{equ:r-constraint}, we formulate the problem into a squared-range least squares (SR-LS) form:
\begin{equation}
	\sum_{i=1}^{N} \| (\mathbf{r}_1\mathbf{p}_i^w + t_x)^2 + (\mathbf{r}_2\mathbf{p}_i^w + t_y)^2 + (\mathbf{r}_3\mathbf{p}_i^w + t_z)^2 - (\mathbf{m}_i)^2 \|^2,
	\label{equ:sr-ls}
\end{equation}
where $\mathbf{r}_k$ is the $k$-th row of $\mathbf{R}$. The only unknown in \eqref{equ:sr-ls} is $t_z$, resulting in a univariate polynomial of degree four. We denote \eqref{equ:sr-ls} by $\mathcal{L}_z$. Examining $\mathcal{L}_z$, the coefficient of $t_z^4$ is always $1$, ensuring the existence of a global minimum of $\mathcal{L}_z$. We find all real roots where $d\mathcal{L}_z/{d t_z} = 0$, check if $d^2\mathcal{L}_z/{d t_z ^2} > 0$, then identify the one that minimizes $\mathcal{L}_z$ as $\hat{t}_z$. The root-finding step is achieved using Matlab command $\href{https://ww2.mathworks.cn/help/matlab/ref/roots.html\#d126e1273729}{\bold{roots}}$ with minimal overhead.

\subsection{Coplanar Cases}
As mentioned in \cite{yang2020extrinsic}, 2D FLS encounters a dual pose ambiguity problem when $\mathbf{p}^w$ are coplanar. This is because if $\mathbf{p}^w$ is mirrored along the imaging plane, exactly the same measurements will be obtained. The authors further note that there is another type of point configuration that might cause dual pose ambiguity, although it is rare in practice. Specifically, if the points in sonar coordinate are symmetric along arbitrary $\theta$ direction, in this case, even if the points are not coplanar, the same effect as mirroring along the imaging plane can be achieved through flipping, thus leading to ambiguity. When this situation occurs, it is sufficient to use one less point to break the symmetry.

Returning to coplanar cases, the ambiguity results in $\dim(\ker(\tilde{\mathbf{Z}})) = 2$ (see Appendix A or \cite{briales2017convex}), which means that the solution $\tilde{\mathbf{r}}^\star$ is expressed as
\begin{equation}
	\tilde{\mathbf{r}}^\star = \alpha_1 \tilde{\mathbf{v}}_1 + \alpha_2 \tilde{\mathbf{v}}_2,
	\label{equ:kernel}
\end{equation}
where $\tilde{\mathbf{r}}_{1:9}^\star$ are the flattened elements of $\mathbf{R}$, $\tilde{\mathbf{r}}_{10}^\star$ is a scaling factor, $\alpha_1$, $\alpha_2$ are unknown coefficients, and $\tilde{\mathbf{v}}_1$, $\tilde{\mathbf{v}}_2$ correspond to the first and second right singular vectors of $\tilde{\mathbf{Z}}$ (see Appendix. A or \cite{briales2017convex}). To determine $\alpha_1$ and $\alpha_2$, we have to utilize the $SO(3)$ constraint. A similar process was mentioned in \cite{wang2024acoustic,zhou2019efficient}, the difference is twofold: 1) In our method, we analyze the null space of a different matrix compared to \cite{wang2024acoustic,zhou2019efficient}. 2) The elements in the null space have a dimension of 10, while in \cite{wang2024acoustic,zhou2019efficient}, the dimension is 9. Since $\alpha_1$ and $\alpha_2$ can implicitly represent the scaling factor, we omit $\tilde{\mathbf{r}}_{10}^\star$ when constructing the $SO(3)$ constraint. The process is summarized in Appendix B for further details.

\subsection{Summary}
For clarity, we summarize the proposed method in Algorithm \ref{alg:pnp_fls}. A constrained iterative optimization (CIO) of \eqref{equ:general-error-func} over all estimated pose parameters is provided as an optional refinement, following the approach proposed in \cite{wang2024acoustic}.
\begin{algorithm}
	\caption{P$n$P in 2D FLS\\
		\textbf{Input:} Paired $\{\mathbf{p}^w_i\in \mathbb{R}^3\}_{i=1}^N$, $\{\mathbf{m}_i\in \mathbb{R}^2\}_{i=1}^N$, optimize\_flag\\
		\textbf{Output:} $\mathbf{R} \in SO(3)$, $\mathbf{t} \in \mathbb{R}^3$}
	\label{alg:pnp_fls}
	\begin{algorithmic}[1]
		\STATE Construct PtL problem as \eqref{equ:cost}
		\STATE Solve PtL problem using convex and global solver (Appendix A, \cite{briales2017convex}), get $\mathbf{R}$ and $\mathbf{t}_{xy}$
		\IF{$\dim(\ker(\tilde{\mathbf{Z}})) = 2$ (coplanar cases)}
		\STATE Find $\alpha_1, \alpha_2$ using $SO(3)$ constraints (Appendix B)
		\STATE Recover valid $\mathbf{R}$
		\ENDIF
		\STATE Solve $t_z$ through \eqref{equ:sr-ls}
		\IF{optimize\_flag = 1}
		\STATE Optimize $\mathbf{R}$, $\mathbf{t}$ over \eqref{equ:general-error-func}
		\ENDIF
		\STATE \textbf{return} $\mathbf{R}$, $\mathbf{t}$
	\end{algorithmic}
\end{algorithm}

\section{Experiments and Results}
\subsection{Experiment Setup}
\textbf{Point Distribution.} We considered both general cases and coplanar cases. The points were generated in full sonar field of view (FoV), specified as $r \in [0,6]$ m, $\theta \in [-30^\circ,30^\circ]$, $\phi \in [-10^\circ,10^\circ]$. For coplanar cases, we forced the plane to pass through point $[0,3,0]^T$ to avoid singular configuration. The dihedral angle between the plane and the $xy$-plane was constrained to be between $5^\circ$ and $70^\circ$, as this represents a reasonable observation angle in practice. Denoting the unit normal of the plane by $\mathbf{n}=[n_x,n_y,n_z]^T$, we imposed the condition $\left(n_y\cdot n_z < 0 \land n_x\cdot n_z > 0\right)$ to avoid the dual-pose problem \cite{yang2020extrinsic}.

\textbf{Point Number.} We evaluated methods performance with [7, 10, 20, 30, 40, 50, 100, 250, 500, 1000] points for general cases, and with [5, 10, 20, $\dots$] points for coplanar cases.

\textbf{Noise Model.} Currently, researchers primarily consider two noise models: 1) \textbf{Polar}: For range-beam sensors, limited angular resolution degrades overall accuracy with increasing range \cite{sheng2024bestanp,su2025rejecting,negahdaripour2005calibration}. 2) \textbf{Cart}: Existing works \cite{yang2020extrinsic,wang2024acoustic,wagner1983statistics,negahdaripour2007integration} claim that Gaussian-distributed errors on the imaging plane ($xy$-axes) translate to Rayleigh-distributed range errors in sonar measurements, matching speckle noise characteristics. Given sonar's extreme noise sensitivity, a comprehensive comparison of model fidelity to real-world conditions exceeds our scope. This work adopted the \textbf{Polar} assumption due to its more intuitive nature. Specifically, we adopted the same value assignment strategy as in \cite{sheng2024bestanp}, where the range noise (in meters) and azimuth angle noise (in radians) were assigned identical values. We tested values in increments of 0.005 below 0.050, except for replacing 0.001 with 0 to avoid singularity issues appeared in method \cite{sheng2024bestanp}. In Figs. \ref{fig:noplanar_varying_points} and \ref{fig:planar_varying_points}, we show the results using 0.025 m and 0.025 rad as baseline noise level to evaluate estimation precision under varying point number. This angular resolution corresponds to $\approx$ 1.43$^\circ$. While advanced ultra-high-frequency (MHz-level) sonar often claims millimeter-level range resolution and sub-degree angular resolution in manuals, real-world applications exhibit significantly higher noise due to the complexity of environment and imperfect knowledge of 3D point positions, whether from measuring artificial markers or triangulation. Moreover, lower-frequency sonars are often preferred in industry for their larger detection range, despite their inherently lower resolution. Thus, 0.025 m, 0.025 rad serves as a realistic approximation of typical noise level in practical scenarios.

\textbf{Metrics.} We randomly selected the coordinates of a point as $\mathbf{t}^{gt}$. For $\mathbf{R}^{gt}$, we randomly generated a quaternion then converted to a rotation matrix. For rotation error metrics, we used $\max^3_{k=1}\arccos(\mathbf{r}_{k}^{gt} \hat{\mathbf{r}}_k^T)/\pi\times180$. For translation, we seperated the evaluation of $\hat{\mathbf{t}}_{xy} = [\hat{t}_x, \hat{t}_y]^T$ and $\hat{t}_z$, following the same as in \cite{wang2024acoustic}, but used $\| \hat{\mathbf{t}}_{xy} - \mathbf{t}_{xy}^{gt} \|$ and $\| \hat{t}_{z} - t_{z}^{gt} \|$ to show the absolute error. Each experiment was run for 300 trials. All simulations were performed on a MacBook Air M1 with 8.0 GB RAM using MATLAB 2024b.

\textbf{Comparison.} We evaluated our original \texttt{PtL} method and its refined version \texttt{PtL-CIO}. Additionally, we replaced the proposed closed-form $t_z$ estimation method with the optimization-based approach from \cite{wang2024acoustic}, termed \texttt{PtL-O$t_z$}, and present the results of its refined version \texttt{PtL-O$t_z$-CIO}. For comparison, we replicated the method proposed in \cite{wang2024acoustic}, named as \texttt{nonapp} (Section IV-B in \cite{wang2024acoustic}), \texttt{app} (Section IV-C in \cite{wang2024acoustic}), and \texttt{CIO} (Section IV-F in \cite{wang2024acoustic}).  The \texttt{CIO} method selects the initial guess of $\mathbf{R}$ and $\mathbf{t}$ from either \texttt{nonapp} or \texttt{app} based on minimum reprojection error, then performs CIO. We have also tested the performance of the recently declared method in \cite{sheng2024bestanp}, called \texttt{BESTAnP}\footnote{https://github.com/LIAS-CUHKSZ/BESTAnP}.

\subsection{General Cases}

\begin{figure*}[!h]
	\centering
	\includegraphics[width=0.32\linewidth]{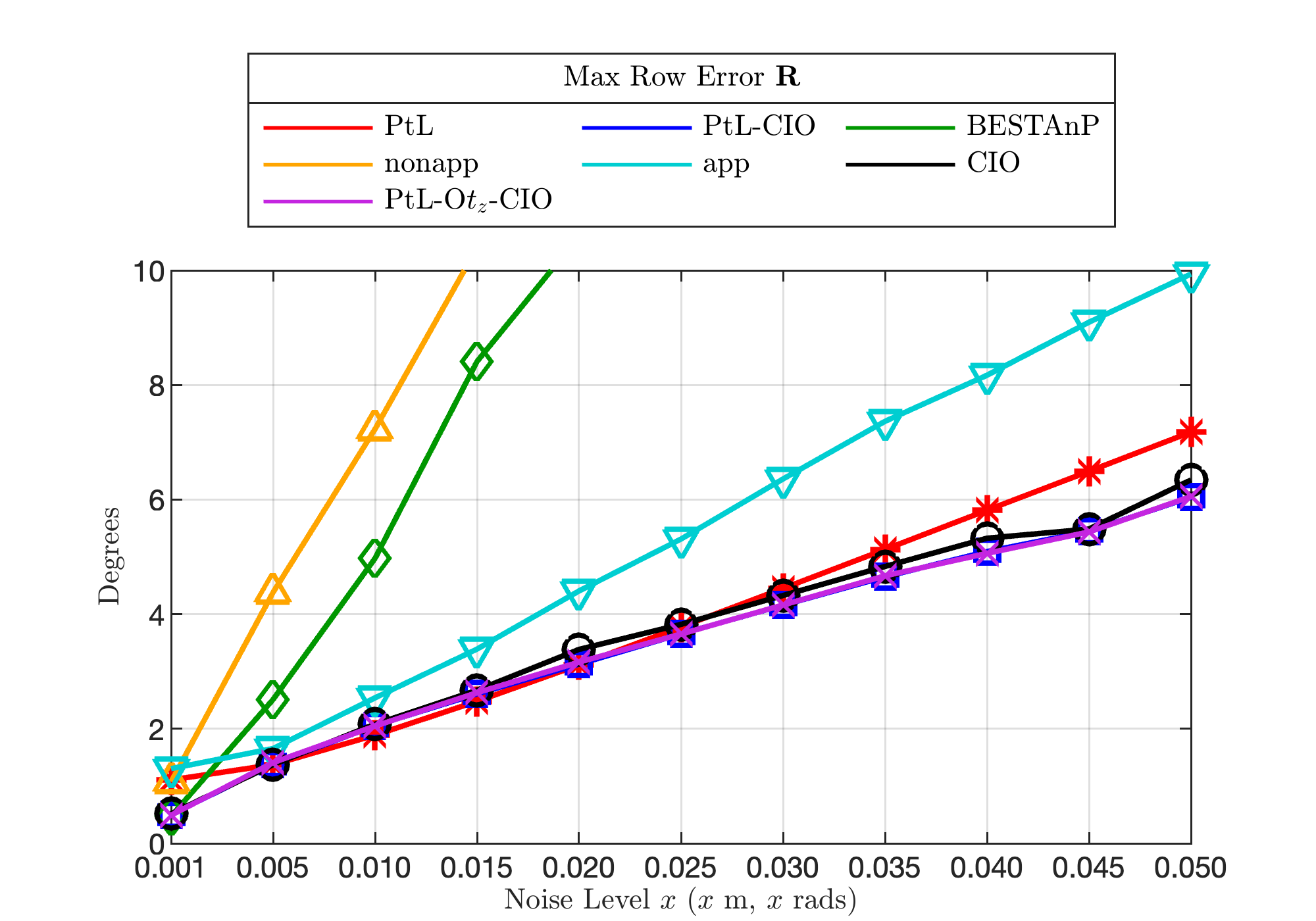}
	\includegraphics[width=0.32\linewidth]{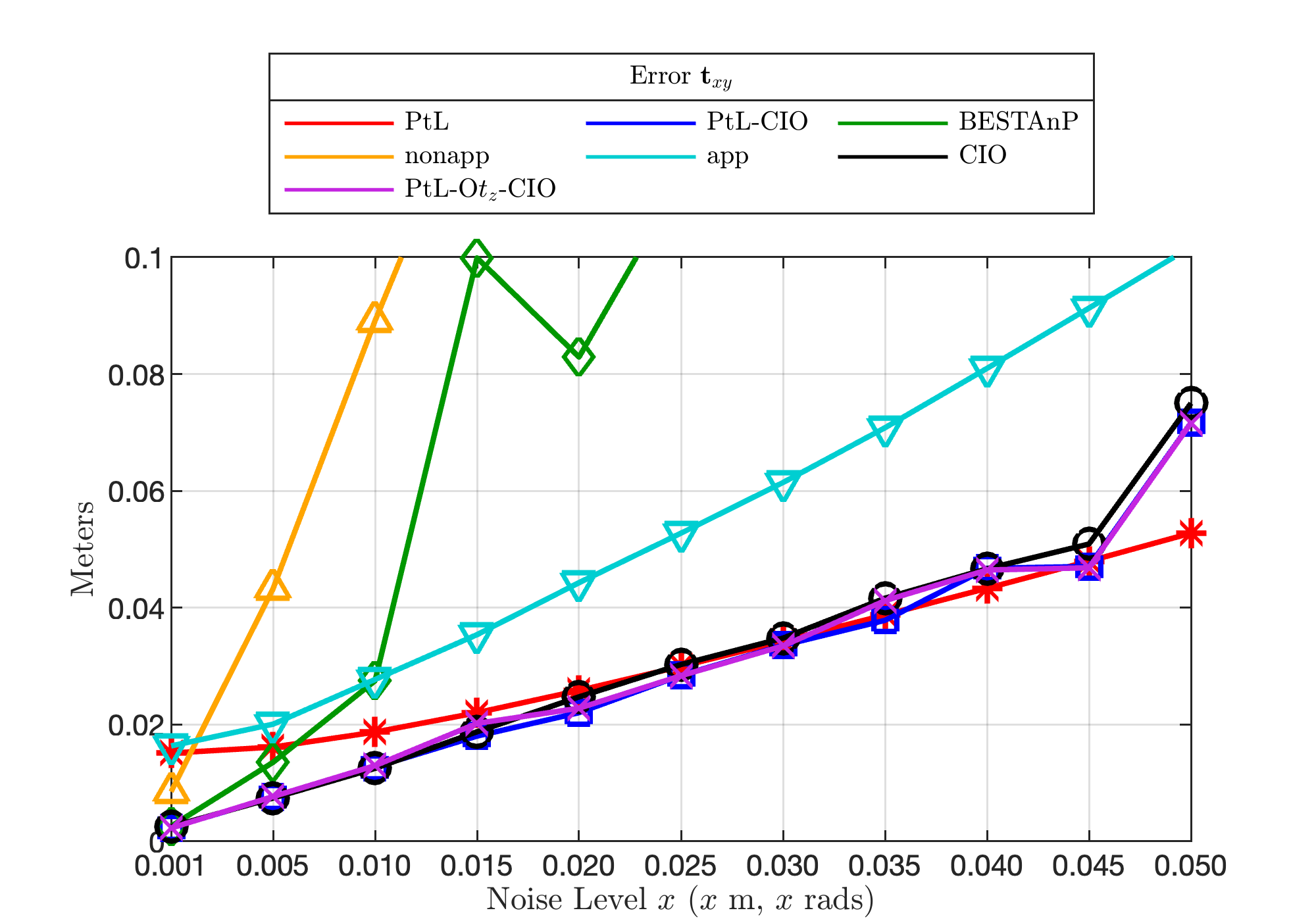}
	\includegraphics[width=0.32\linewidth]{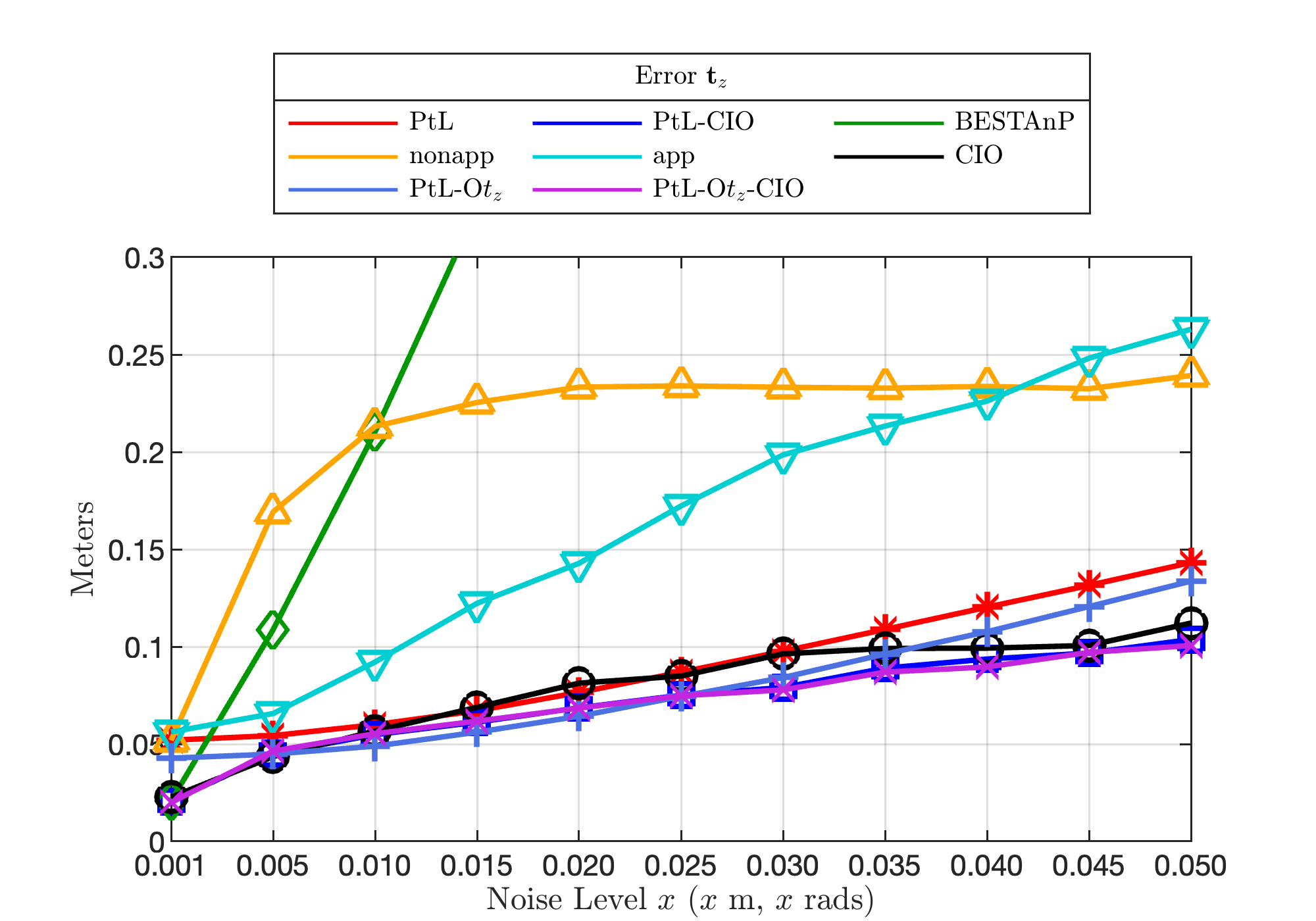}
	\caption{Results for the general cases with increasing noise level under 20 points. From left to right: angular error, $\mathbf{t}_{xy}$ error, and $t_z$ error.}
	\label{fig:noplanar_varying_noise}
\end{figure*}

The results for general cases with increasing noise levels under 20 points are shown in Fig. \ref{fig:noplanar_varying_noise}. As noise increases, all methods exhibit performance degradation. The \texttt{nonapp} method, while theoretically exact due to its approximation-free formulation, suffers significant accuracy loss as noise increases because it relies solely on angular measurements. In \texttt{BESTAnP}, the observability of $t_z$ is low. When noise increases, its impact on the overall estimation becomes substantial. While \texttt{app} maintains relatively stable performance, its orthographic approximation introduces noticeable bias in low-noise cases. Our proposed \texttt{PtL} method similarly shows some bias at low noise levels, but achieves progressively clearer advantages as noise increases. For rotation estimation, \texttt{PtL} directly searches the $SO(3)$ manifold to guarantee optimality, unlike \texttt{app} which first computes an affine matrix and then projects it onto $SO(3)$. For $\mathbf{t}_{xy}$, \texttt{PtL} incorporates information from all available points while \texttt{app} arbitrarily selects a single reference point. In estimating $t_z$, \texttt{PtL} lacks the constraints as employed by \cite{wang2024acoustic} and \eqref{equ:general-error-func}, the superior performance in the other five degrees of freedom creates more favorable conditions for the proposed closed-form $t_z$ solver to produce accurate results. If applying the optimization-based solver from \cite{wang2024acoustic}, further accuracy improvements can be achieved as demonstrated by \texttt{PtL-O$t_z$}. Now considering the additional optimization-based refinement over reprojection residual, all the methods show similar performance. The \texttt{CIO} shows significant improvement. If the refinement starts from \texttt{PtL}, the improvements are limited, but also show comparable or slightly better results than \texttt{CIO}. The slight superiority may come from better starting point. We would like to note that sometimes \texttt{PtL} even performs better than refinement-based method.

\begin{figure*}[!h]
	\centering
	\includegraphics[width=0.32\linewidth]{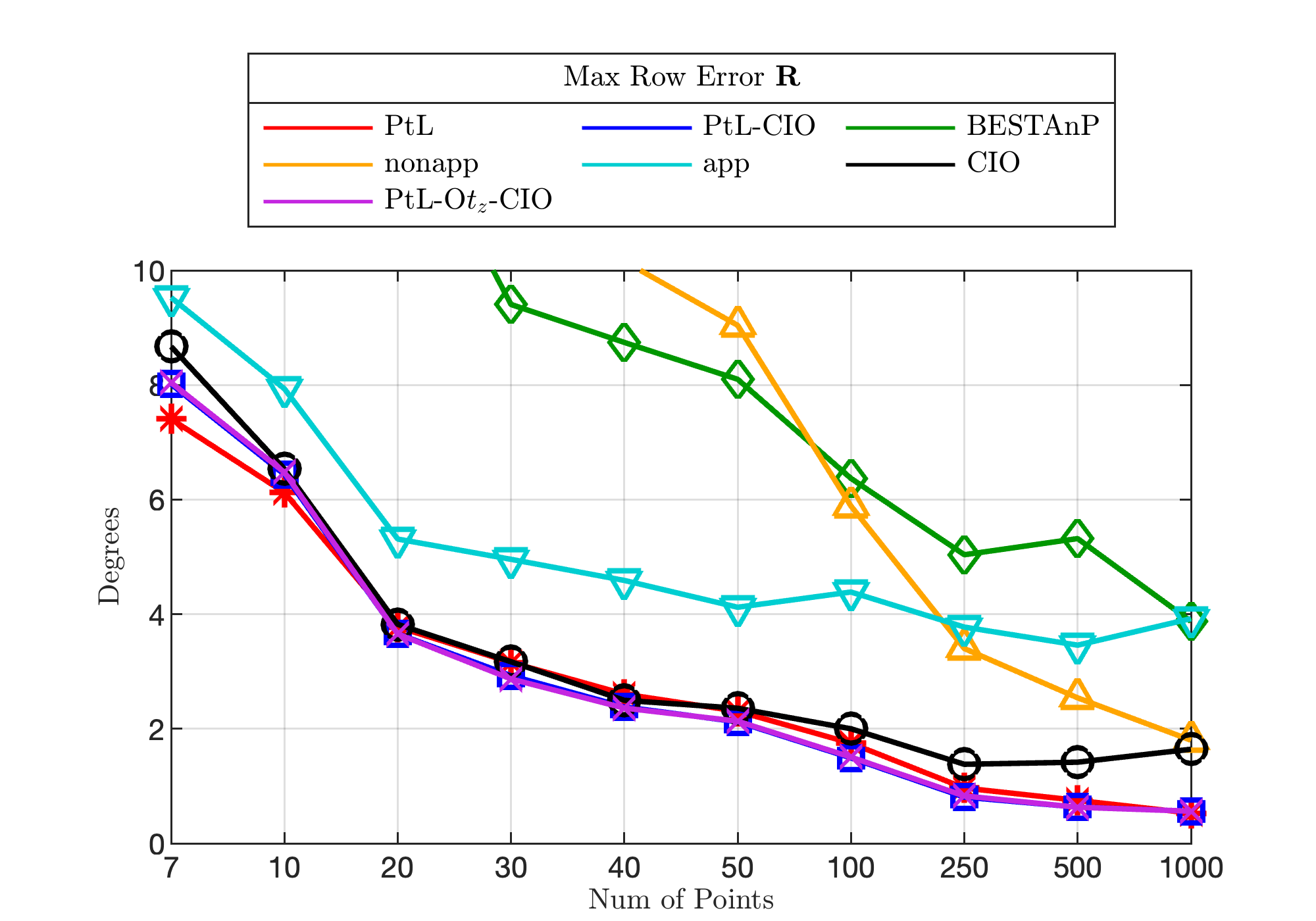}
	\includegraphics[width=0.32\linewidth]{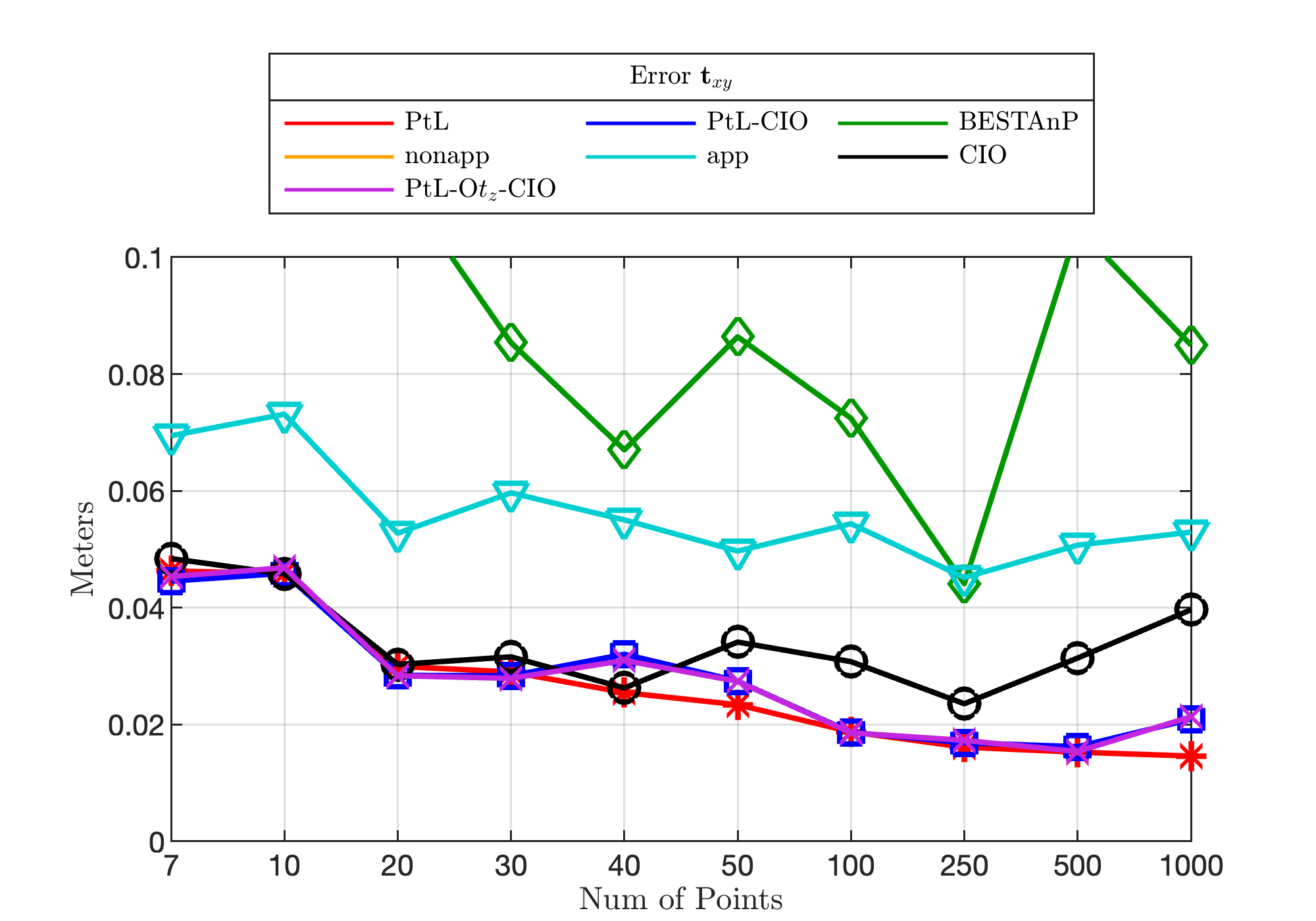}
	\includegraphics[width=0.32\linewidth]{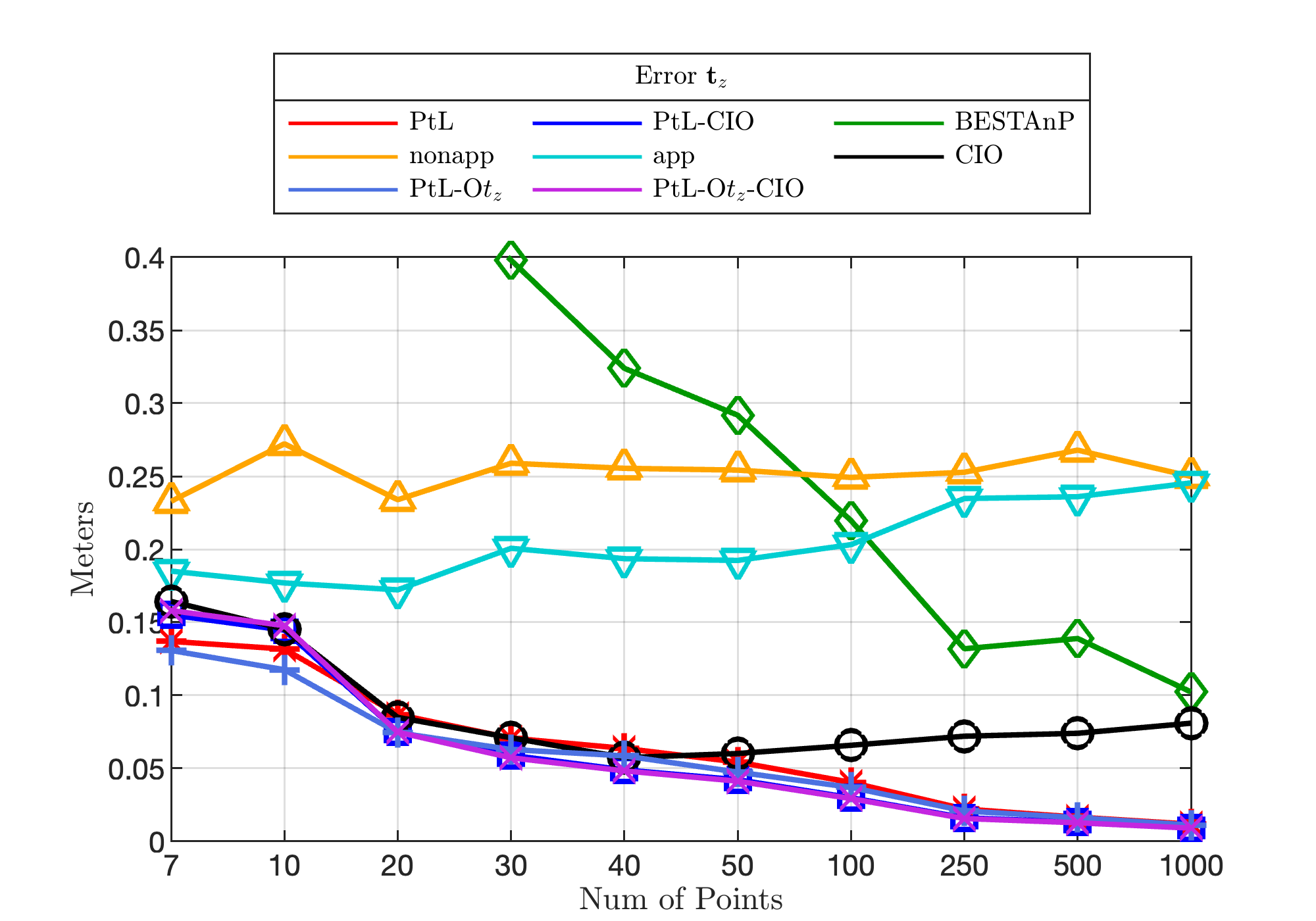}
	\caption{Results for the general cases with increasing point number under 0.025 m, 0.025 rad noise. From left to right: angular error, $\mathbf{t}_{xy}$ error, and $t_z$ error.}
	\label{fig:noplanar_varying_points}
\end{figure*}

The results for general cases with increasing point number under 0.025 m, 0.025 rad noise are shown in Fig. \ref{fig:noplanar_varying_points}. When the number goes large, the performance of all the methods improves. The performance differences between methods are similar to aforementioned analysis. We would like to note the more evident improvements of the proposed method at both low (7) and super high (1000) point numbers. It shows better stability near minimal cases, and the different convergence towards a more accurate estimate highlights the impact of better initial guesses compared to \cite{wang2024acoustic}.

\subsection{Coplanar Cases}

\begin{figure*}[!h]
	\centering
	\includegraphics[width=0.32\linewidth]{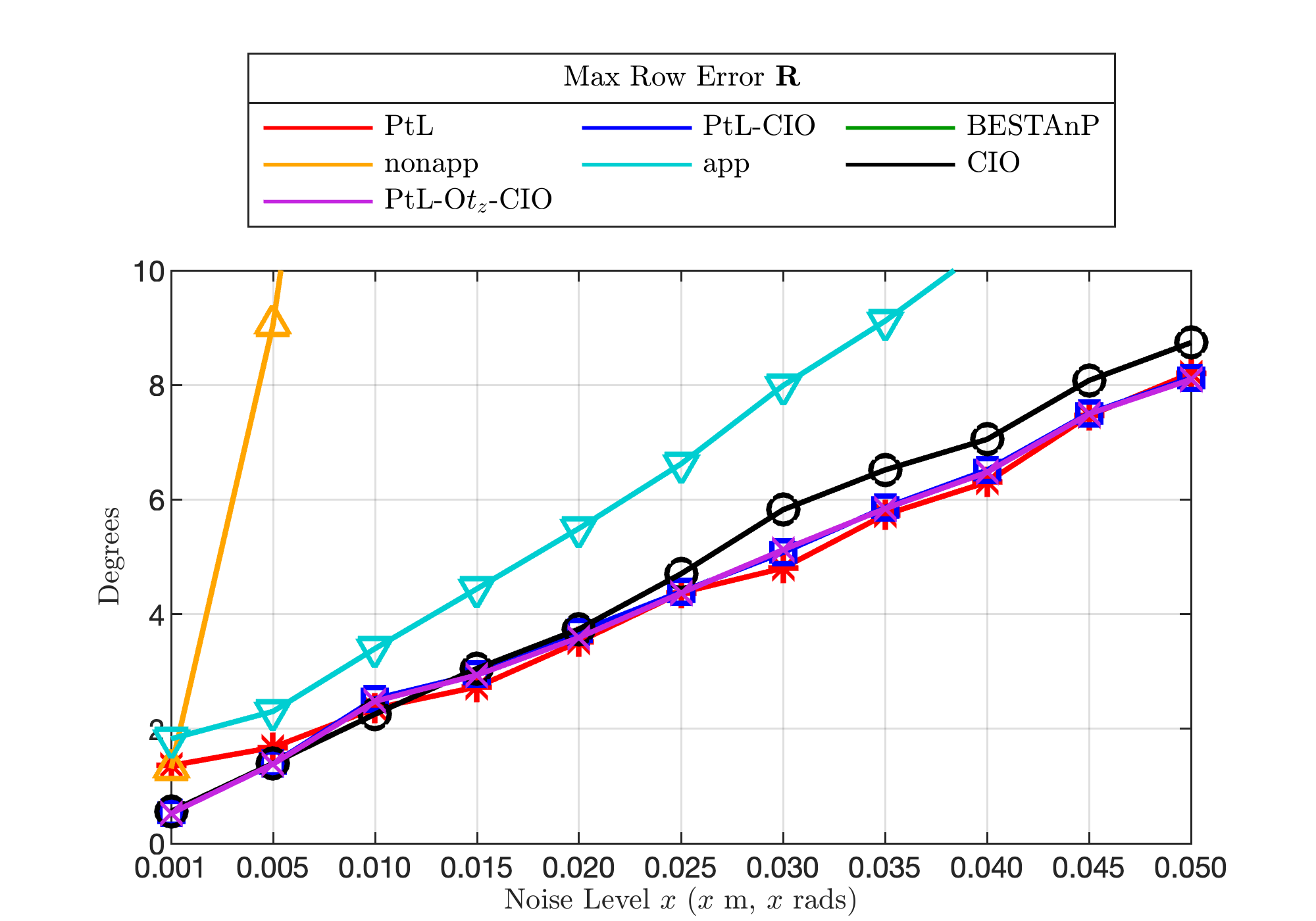}
	\includegraphics[width=0.32\linewidth]{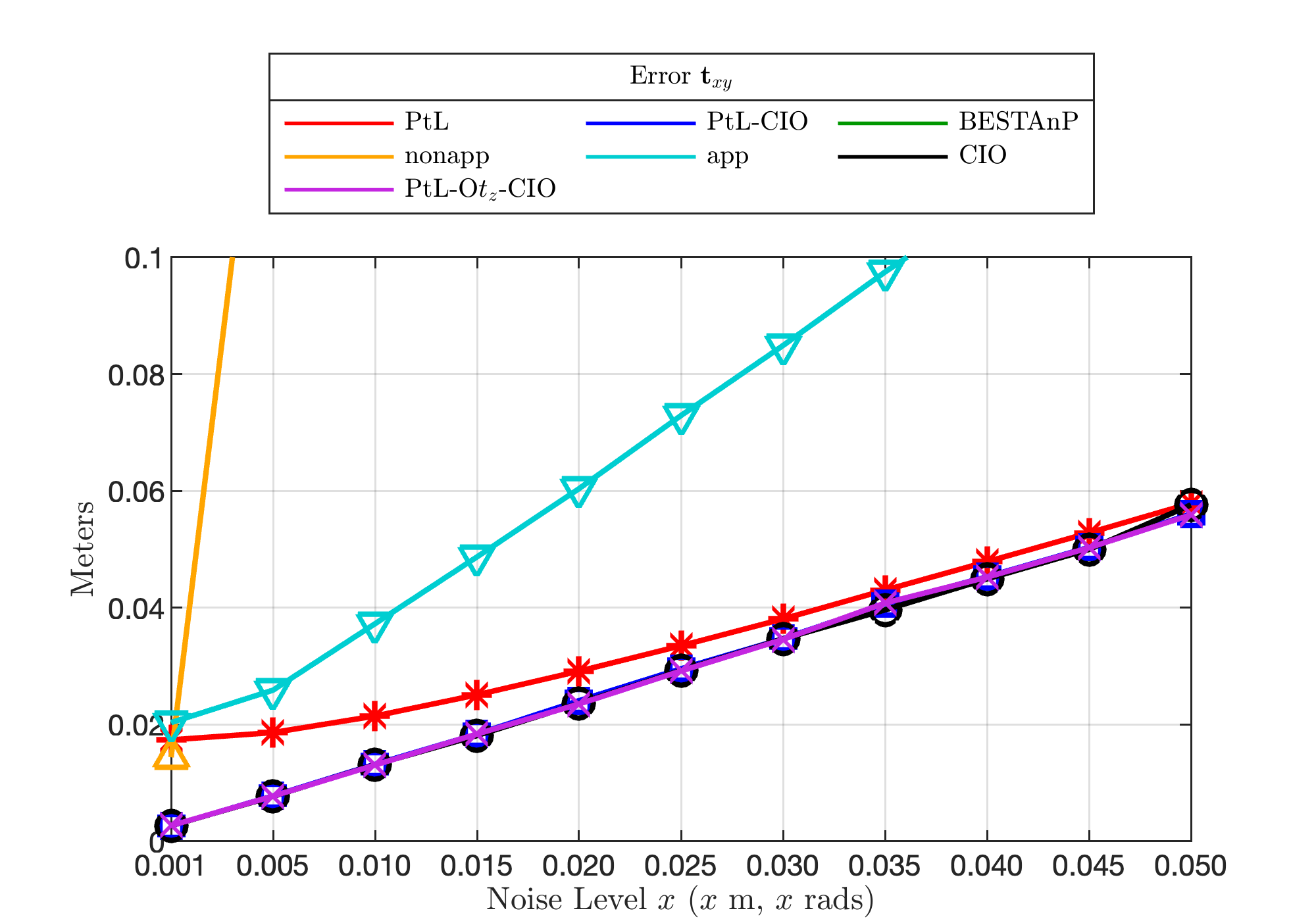}
	\includegraphics[width=0.32\linewidth]{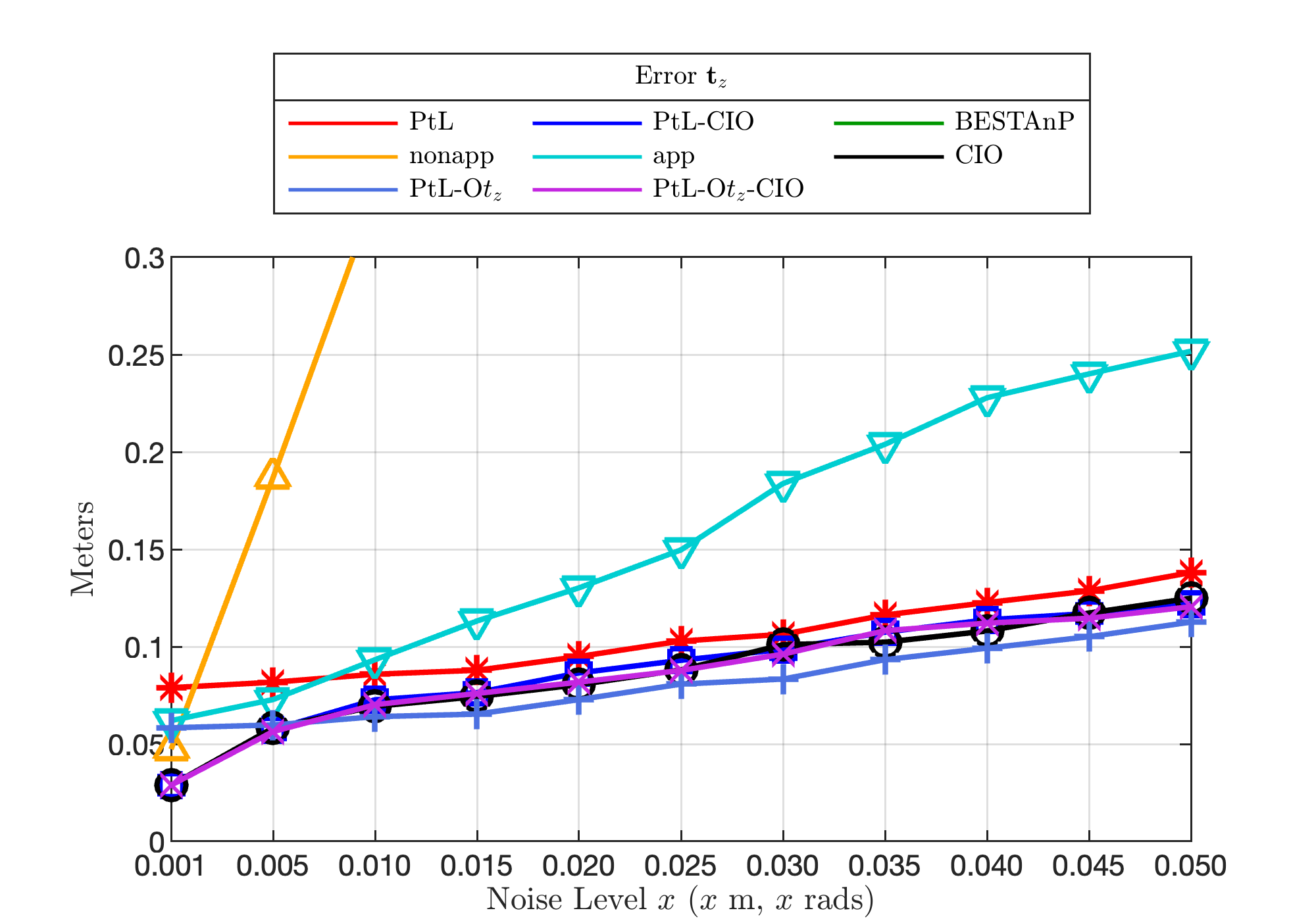}
	\caption{Results for the coplanar cases with increasing noise level under 20 points. From left to right: angular error, $\mathbf{t}_{xy}$ error, and $t_z$ error.}
	\label{fig:planar_varying_noise}
\end{figure*}

\begin{figure*}[!h]
	\centering
	\includegraphics[width=0.32\linewidth]{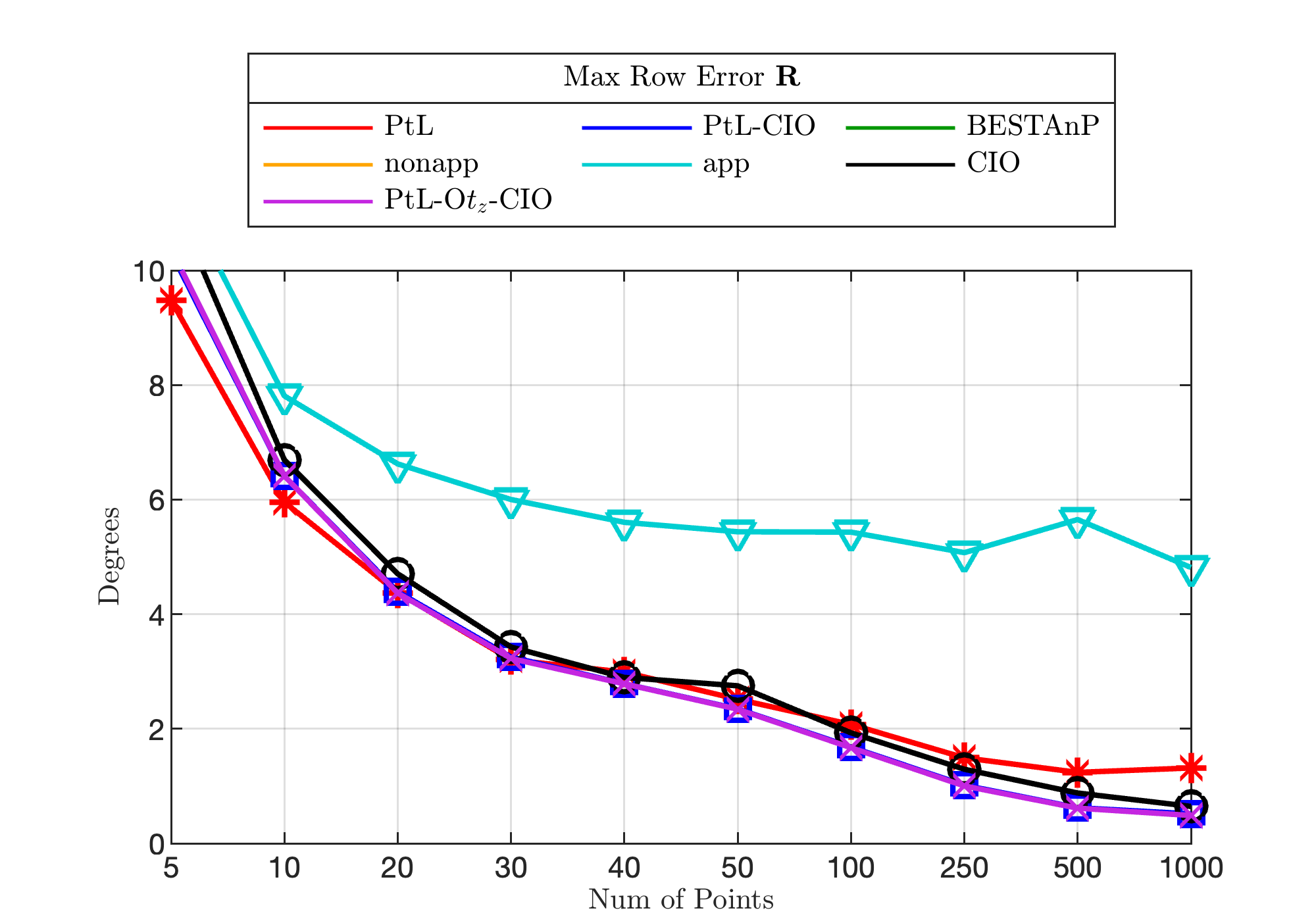}
	\includegraphics[width=0.32\linewidth]{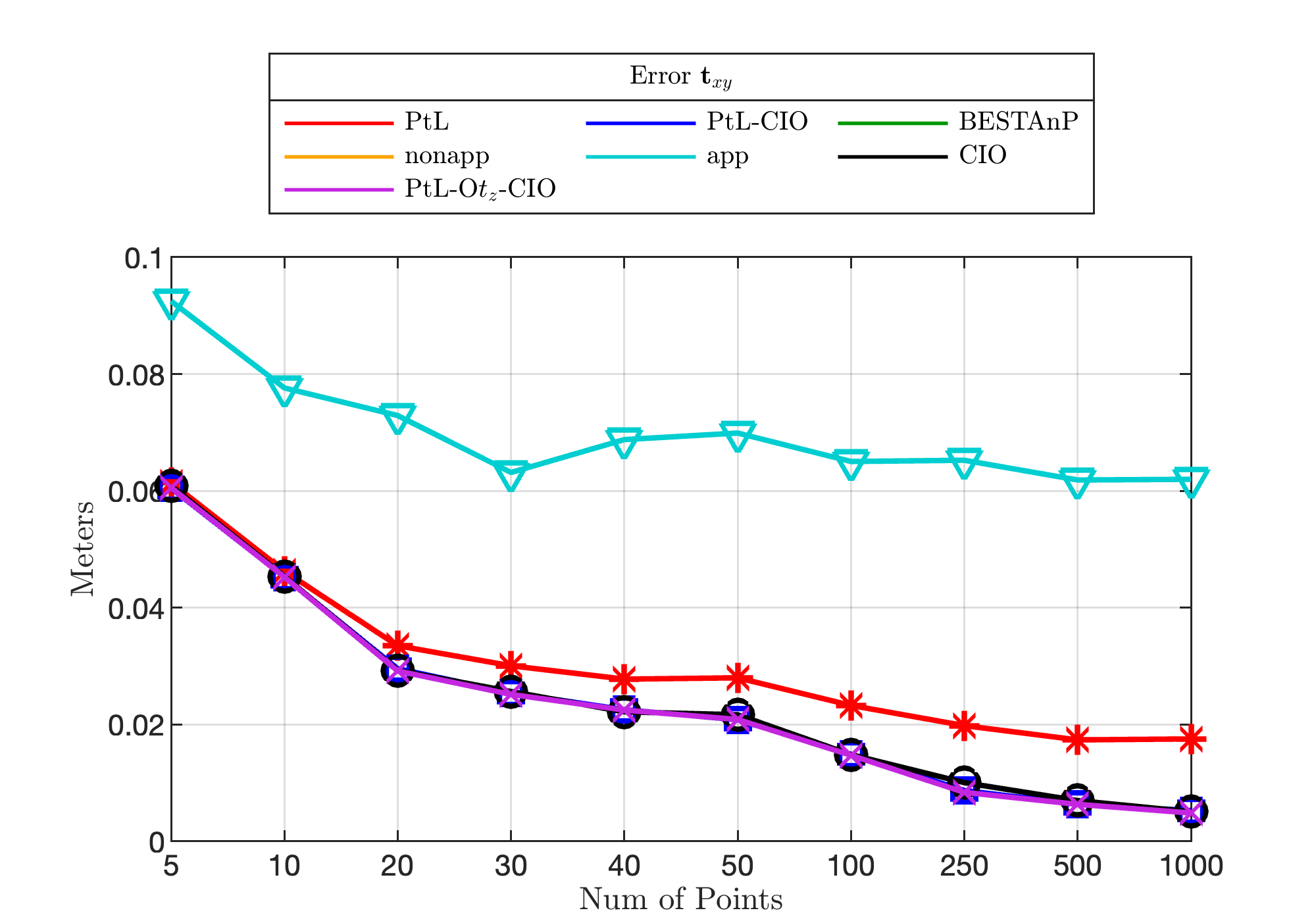}
	\includegraphics[width=0.32\linewidth]{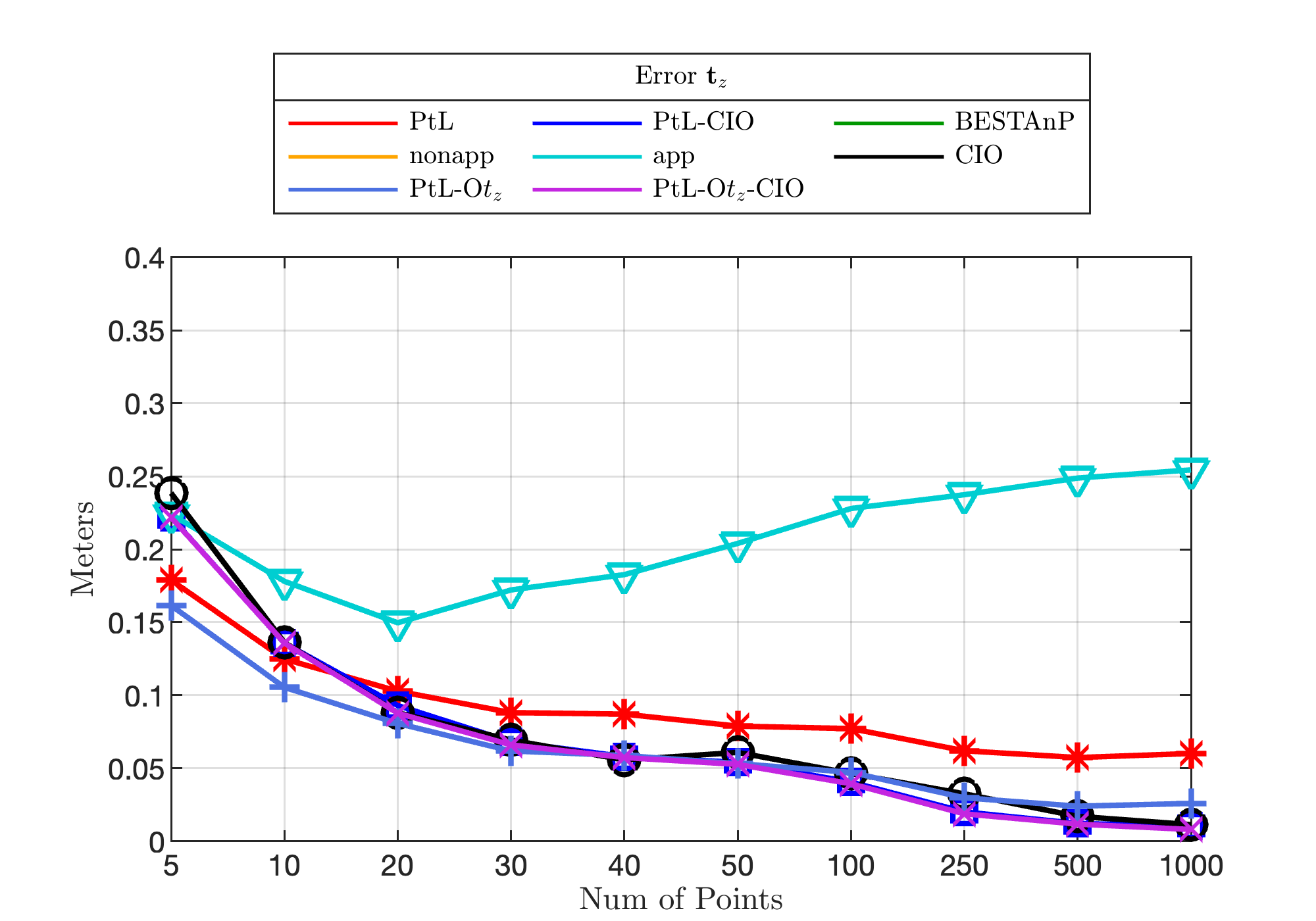}
	\caption{Results for the coplanar cases with increasing point number under 0.025 m, 0.025 rad noise. From left to right: angular error, $\mathbf{t}_{xy}$ error, and $t_z$ error.}
	\label{fig:planar_varying_points}
\end{figure*}

The results are shown in Figs. \ref{fig:planar_varying_noise} and \ref{fig:planar_varying_points}. For \texttt{BESTAnP}, it cannot handle coplanar conditions, and the error is extremely high. The overall difference between the methods is similar to that observed in the general cases, except that for the translation estimate, \texttt{PtL} performs slightly worse than in the general cases, but it is still significantly better than other initial guesses.

\subsection{Time Costs}
\begin{table*}[htbp]
	\centering
	\caption{The results of the average time costs in milliseconds when running with 20 points.}
	\label{tab:time_costs}
	\begin{tabular}{cccccccccc}
		\toprule
		& \texttt{PtL} & \texttt{BESTAnP} & \texttt{nonapp} & \texttt{app}  & \texttt{Closed-Form $t_z$} & \texttt{Optimize $t_z$} & \texttt{Optimize 6 DoF} & \texttt{PtL-CIO} & \texttt{CIO} \\
		\midrule
		General (ms) & 352.57 & 3.09 & 222.52 & 91.23 & 0.48 & 14.32 & 22.63 & 375.20 & 336.38 \\
		Coplanar (ms) & 672.57 & 4.40 & 24.53 & 8.84 & 0.35 & 10.06 & 9.90 & 682.47 & 43.27\\
		\bottomrule
	\end{tabular}
\end{table*}
Tab. \ref{tab:time_costs} shows the average time costs of different methods with 20 points under general and coplanar cases. For \texttt{Optimize $t_z$}, the time cost represents the duration solely for estimating the $t_z$ parameter through optimization, not the entire runtime of a complete method. The same interpretation applies to \texttt{Closed-Form $t_z$} and \texttt{Optimize 6 DoF}. Among all the methods, \texttt{BESTAnP} is the fastest. In general cases, \texttt{nonapp} and \texttt{app} have to conduct null space analysis, which incurs a higher time cost. The null space analysis is not fully optimized for speed in our implementation, and the time cost can be further compressed. The optimization process takes around 10-20 ms, while the overall time cost for \texttt{CIO} is 300+ ms. The proposed \texttt{PtL} is slightly slower than \texttt{CIO}. For coplanar cases, \texttt{nonapp} and \texttt{app} do not require null space analysis, resulting in a significant reduction in time cost. However, \texttt{PtL} requires null space analysis at this time, which increases its overall time cost in coplanar cases. In Table \ref{tab:time_costs}, the results also show that our closed-form $t_z$ solver is quite efficient compared to optimization-based methods.

\section{Conclusions}

In this paper, we proposed to solve the P$n$P problem in 2D FLS within a 3D PtL paradigm. The results demonstrate significant improvements over non-reprojection-optimized methods. When followed by an optimization-based refinement, further improvements can be achieved. The main drawbacks of the proposed method are: 1) Unconstrained $t_z$ estimation may lead to a performance decrease when noise is high. 2) The time cost is too high for real-time robotic applications. In future work, we would like to adopt advanced closed-form PtL solvers to test the precision of the estimate and check if they can retrieve solutions in degenerate scenarios, i.e. coplanar cases.

\renewcommand{\theequation}{A\arabic{equation}} 
\setcounter{figure}{0}
\renewcommand{\thefigure}{A\arabic{figure}}
\setcounter{equation}{0} 
\section*{Appendix}

\subsection{Duality-Based Optimal Solver}
In \cite{briales2017convex}, the registration problem was formulated as a quadratically constrained quadratic program (QCQP):
\begin{subequations}
	\begin{align}
		& \min_\mathbf{R} && \tilde{\mathbf{r}}^T\tilde{\mathbf{Q}} \tilde{\mathbf{r}}, \quad \tilde{\mathbf{r}} = [\mathbf{c_1}^T, \mathbf{c_2}^T, \mathbf{c_3}^T, h]^T,
		\label{equ:primal0}
		\\
		& \text{s.t.} && \mathbf{R}^T\mathbf{R} = h^2\mathbf{I}_3,
		\label{equ:primal1}
		\\
		& && \mathbf{R}\mathbf{R}^T = h^2\mathbf{I}_3,
		\label{equ:primal2}
		\\
		&   && \mathbf{r}^T_{k_1} \times \mathbf{r}^T_{k_2} = h\mathbf{r}^T_{k_3}, {k_1},{k_2},{k_3} = \text{cyclic}(1,2,3),
		\label{equ:primal3}
		\\
		& && h^2 = 1.
	\end{align}
	\label{equ:primal}
\end{subequations}
In \eqref{equ:primal}, $\mathbf{r}_{\{1,2,3\}}, \mathbf{c}_{\{1,2,3\}}$ are the rows and the columns of $\mathbf{R} \in SO(3)$, respectively. The $\tilde{\mathbf{Q}}$ in \eqref{equ:primal0} is the coefficient matrix of the registration problem, $h$ is the homogeneous variable. The \eqref{equ:primal1} and \eqref{equ:primal2} indicate the orthonormality of $\mathbf{R}$. For the determinant constraint $\det(\mathbf{R}) = 1$, a right-hand rule is used instead \cite{tron2015inclusion}, forming quadratic constraints as in \eqref{equ:primal3}. This formulation marginalizes out $\mathbf{t}$ in the original problem. Through duality theory, the QCQP is transformed into a small semidefinite program (SDP):
\begin{equation}
	d^\star = \max_{\tilde{\boldsymbol{\lambda}}} \gamma, \quad \text{s.t.} \ \tilde{\mathbf{Z}}(\tilde{\boldsymbol{\lambda}}) \succeq \mathbf{0},
\end{equation}
where the $\tilde{\boldsymbol{\lambda}} = [\boldsymbol{\lambda}^T, \gamma]^T$ is a vector gathers the dual variables to all the constraints in \eqref{equ:primal}. The solution of $\tilde{\mathbf{r}}^\star$ lies in the null space of $\tilde{\mathbf{Z}}$, where $\dim(\ker(\tilde{\mathbf{Z}})) = 1$, implying that the solution is recovered up to a scale factor. The detailed forms of $\tilde{\mathbf{Q}}$ and $\tilde{\mathbf{Z}}$, as well as the process of retrieving $\mathbf{t}$ are provided in \cite{briales2017convex}.

\subsection{Solving for $\alpha_1$, $\alpha_2$}

Using the $SO(3)$ constraint, $\alpha_1$ and $\alpha_2$ can be represented as $\mathbf{Fa}=\mathbf{b}$, where $\mathbf{a}=[\alpha_1^2,\alpha_2^2,\alpha_1\alpha_2]$, $\mathbf{b}=[0,\dots,1]$, with $0$ representing orthogonality constraints and $1$ representing the normalization constraint. If we treat $\alpha_1^2,\alpha_2^2,\alpha_1\alpha_2$ as independent variables, a quick solution can be obtained using the pseudo-inverse of $\mathbf{F}$. However, this linearization method neglects the relationships between the variables and may yield suboptimal solutions. Therefore, we transform the problem into a least squares form:
\begin{equation}
	\mathop{\arg \min}_{\alpha_1,\alpha_2} M(\alpha_1,\alpha_2), \quad M(\alpha_1, \alpha_2) = \| \mathbf{Fa} - \mathbf{b} \|^2_2.
\end{equation}
Seeking its first-order optimality conditions:
\begin{equation}
	g_1=\frac{\partial M}{\partial \alpha_1} = 0, \quad g_2=\frac{\partial M}{\partial \alpha_2} = 0.
	\label{equ:dual:M_optimal}
\end{equation}
This problem involves solving a system of bivariate polynomial equations of degree up to 3, with monomials $[\alpha_1^3,\alpha_1^2\alpha_2,\alpha_1,\alpha_2^2,\alpha_1,\alpha_2]$. To address this, the algorithm employs the hidden variable method. In this method, one of the unknowns is treated as a constant. Assuming $\alpha_1$ is the unknown and $\alpha_2$ is the hidden variable, we have:
\begin{equation}
	g_i = c_{i1}\alpha_1^3 + c_{i2}(\alpha_2)\alpha_1^2 + c_{i3}(\alpha_2)\alpha_1 + c_{i4}(\alpha_2) = 0,
\end{equation}
where $c_{ij}(\alpha_2)$ are polynomials of $\alpha_2$. For $g_1$ and $g_2$ to hold true simultaneously, they must have a common root, which means their resultant must be equal to 0. Denoting $c_{ij}(\alpha_2)$ as $c_{ij}$, we have:
\begin{equation}
	\det \begin{pmatrix}
		c_{11} & 0 & 0 & c_{21} & 0 & 0 \\
		c_{12} & c_{11} & 0 & c_{22} & c_{21} & 0 \\
		c_{13} & c_{12} & c_{11} & c_{23} & c_{22} & c_{21} \\
		c_{14} & c_{13} & c_{12} & c_{24} & c_{23} & c_{22} \\
		0 & c_{14} & c_{13} & 0 & c_{24} & c_{23} \\
		0 & 0 & c_{14} & 0 & 0 & c_{24}
	\end{pmatrix} = 0.
	\label{equ:dual:resultant}
\end{equation}
The determinant on the left side of \eqref{equ:dual:resultant} is the resultant of $g_1$ and $g_2$. Expanding \eqref{equ:dual:resultant} yields a univariate polynomial of degree up to 9:
\begin{equation}
	\sum_{i=0}^{9}k_i \alpha_2^{i}=0.
	\label{equ:dual:9th}
\end{equation}
$\alpha_2$ can be obtained through numerical methods, and substituting it back into $g_1$ and $g_2$ yields $\alpha_1$. It is important to note that solving \eqref{equ:dual:9th} will yield multiple solutions, resulting in multiple solutions for $\alpha_1$. Each set of solutions is substituted back into $M$ for verification, and the one that yields the minimum value is selected as the output. Through the above analysis, the case of coplanar landmarks can be handled.

\bibliographystyle{IEEEtran}
\bibliography{ref}

\addtolength{\textheight}{-12cm}

\end{document}